\pdfoutput=1

\documentclass[11pt]{article}

\usepackage[preprint]{acl}

\usepackage{times}
\usepackage{latexsym}

\usepackage[T1]{fontenc}

\usepackage[utf8]{inputenc}

\usepackage{microtype}

\usepackage{inconsolata}

\usepackage{graphicx}
\usepackage{amsmath}

\title{Knowledge Graphs are all you need: Leveraging KGs in Physics Question Answering}

\author{Krishnasai Addala \\
  IIIT Delhi \\
  Delhi, India \\
  \scalebox{0.83}{\texttt{krishnasai20442@iiitd.ac.in}} \\\And
  Kabir Dev Paul Baghel \\
  IIIT Delhi \\
  Delhi, India \\
  \scalebox{0.83}{\texttt{kabir20564@iiitd.ac.in}} \\\And
  Dhruv Jain \\
  IIT BHU \\
  Uttar Pradesh, India \\
  \scalebox{0.83}{\texttt{dhruv.jain.ece21@itbhu.ac.in}} \\\AND
  Navya Gupta \\
  MIDAS Lab \\
  Delhi, India \\
  \scalebox{0.90}{\texttt{guptanavya1808@gmail.com}} \\\And
  Rishitej Reddy Vyalla \\
  IIIT Delhi \\
  Delhi, India \\
  \scalebox{0.90}{\texttt{rishitej23439@iiitd.ac.in}} \\\And
  Chhavi Kirtani \\
  IIIT Delhi \\
  Delhi, India \\
  \scalebox{0.83}{\texttt{chhavi18229@iiitd.ac.in}} \\\AND
  Avinash Anand \\
  IIIT Delhi \\
  Delhi, India \\
  \scalebox{0.83}{\texttt{avinasha@iiitd.ac.in}} \\\And
  Rajiv Ratn Shah \\
  IIIT Delhi \\
  Delhi, India \\
  \scalebox{0.83}{\texttt{rajivratn@iiitd.ac.in}} \\
  }

\begin{document}
\maketitle
\vspace*{3cm}
\begin{abstract}

This study explores the effectiveness of using knowledge graphs generated by large language models to decompose high school-level physics questions into sub-questions. We introduce a pipeline aimed at enhancing model response quality for Question Answering tasks. By employing LLMs to construct knowledge graphs that capture the internal logic of the questions, these graphs then guide the generation of sub-questions. We hypothesize that this method yields sub-questions that are more logically consistent with the original questions compared to traditional decomposition techniques. Our results show that sub-questions derived from knowledge graphs exhibit significantly improved fidelity to the original questions' logic. This approach not only enhances the learning experience by providing clearer and more contextually appropriate sub-questions but also highlights the potential of LLMs to transform educational methodologies. The findings indicate a promising direction for applying AI to improve the quality and effectiveness of educational content.
\end{abstract}



\section{Introduction}

In recent years, Question answering and reasoning have emerged as a crucial application of Large Language Models (LLMs). LLMs have shown remarkable performance in processing and generating human-like responses and reasoning to variety of questions. Recent research has rapidly advanced reasoning capabilities of LLMs in diverse domains, driven by innovations in retrieval-augmented generation \cite{anand2023kg, anand2023sciphyrag}, multimodal\\ \\ \\ \\ \\ \\ \\ \\ \\learning \cite{anand2024mm} ,  and specialized datasets   \cite{anand2024mathify, anand2024geovqa}. Such advancements underline the flexibility of LLMs to adapt to diverse  real-world tasks requiring nuanced reasoning. However, their ability to handle complex, multi-step questions that require logical reasoning and domain-specific knowledge remains a significant challenge. One promising approach to enhance LLM's QA capabilities is to leverage knowledge graphs (KGs)~\cite{zhang2022embedding,wang2021keytextretrieval}. KGs provide a structured representation of information, capturing entities, relationships, and their properties. By integrating KGs into the question-answering process, LLMs can benefit from a more organized and contextually relevant knowledge base, potentially leading to improved reasoning and answer quality.

In the domain of education, particularly in subjects like physics, question-answering often involves convoluted problem-solving that requires a thorough grasp of foundational principles and the ability of decomposing complex problems into simpler, actionable components. Traditional question decomposition techniques~\cite{perez2020unsupervised,xu2022decomprc} have shown limited success in generating sub-questions that maintain logical consistency and coherence with the original question.

This study proposes a novel approach that combines the power of LLMs with knowledge graphs to improve question answering in the context of high school-level physics problems. We introduce a pipeline that utilizes LLMs to generate knowledge graphs from the original question, capturing its internal logic and key relationships. These knowledge graphs are then used to guide the model in generating sub-questions that are more faithful to the original question's intent and structure.
\begin{figure}[h!]
  \centering
  \includegraphics[width=0.5\textwidth]{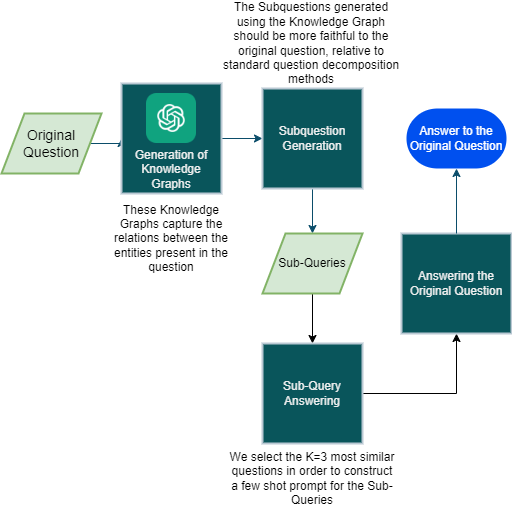}
  \caption{Overview of the proposed QA pipeline: The original question is transformed into a Knowledge Graph to generate sub-questions. Answers to these sub-questions are then used to guide the response to the original question.}
  \label{fig:image2}
\end{figure}
We hypothesize that this knowledge graph-based question decomposition approach will lead to sub-questions that are more logically consistent and relevant to the original question, ultimately resulting in higher-quality answers. By providing clearer and contextually appropriate sub-questions, this method aims to enhance the learning experience and demonstrate the potential of LLMs in revolutionizing educational methodologies.

The major contributions of this study are given below:
\begin{enumerate}
\item A novel pipeline that integrates knowledge graphs with LLMs for improved question answering in high school-level physics.
\item An investigation of the effectiveness of knowledge graph-based question decomposition in generating logically consistent and relevant sub-questions.
\item A new dataset comprising high-quality, high-school physics questions, augmented with corresponding internal knowledge graphs and subqueries generated using advanced models such as Gemini Pro. This dataset serves as a unique resource for fine-tuning open-source LLMs to replicate and expand upon the sophisticated analytical abilities demonstrated by larger models.
\item Insights into the potential of LLMs in enhancing the quality and effectiveness of educational content.
\end{enumerate}
The structure of this paper is as follows: Section 2 presents a review of related work in the areas of reasoning, question answering, question decomposition, and the use of knowledge graphs in LLMs. Section 3 describes the dataset used in this study~\cite{anand2023revolutionizing}, while Section 4 details the proposed methodology. Section 5 presents the experimental results and analysis, followed by Section 6 that discusses the limitations and outlines potential future directions, and Section 7 concludes the paper.

\section{Related Work}
\paragraph{Reasoning in LLMs} 
By combining contextual knowledge and domain-specific training, LLMs are increasingly able to handle complex reasoning tasks with higher accuracy and robustness. In scientific reasoning, methods like knowledge graph-guided citation generation (KG-CTG) \cite{anand2023kg} have improved the ability of LLMs to generate accurate and contextually appropriate scientific citations. These approaches leverage external knowledge sources and multi-source learning to enhance the relevance and coherence of generated text in academic domains. In mathematical and physical reasoning, retrieval-augmented techniques and multimodal prompting have led to significant performance improvements. Studies such as Sciphyrag \cite{anand2023sciphyrag} and MM-PhyQA \cite{anand2024mm} demonstrate how retrieval-based models and multi-image chain-of-thought prompting can enhance reasoning on physics and STEM-related questions. Similarly, the Mathify \cite{anand2024mathify} framework evaluates LLMs on mathematical problem-solving tasks, contributing to systematic improvements in mathematical reasoning . The use of domain-specific datasets \cite{anand2024geovqa, anand2024mathify} has further strengthened LLM reasoning capabilities in education and multimodal learning. 
\paragraph{Question Answering in LLMs} LLMs have demonstrated impressive performance on various question-answering tasks, leveraging their ability to process and reason over natural language inputs. However, their performance can still be limited by the lack of structured knowledge representations and reasoning capabilities required for complex question-answering scenarios \cite{yang2022investigating, petrochuk2022nqa}. Recent work has explored methods to enhance LLM question-answering abilities, such as retrieval-augmented generation \cite{lewis2020retrievalqna}, multi-hop reasoning \cite{zhu2022exhibiting}, and incorporating external knowledge sources \cite{wang2022kpal}.

\paragraph{Question Decomposition} Breaking down complex questions into simpler sub-questions, known as question decomposition, has been widely studied in natural language processing (NLP) \cite{khashabi2018question, jiang2019self}. Traditional approaches relied on rule-based systems or statistical methods, often producing sub-questions lacking contextual coherence or logical consistency with the original question. Recent work has explored neural networks and language models for generating more coherent and logically consistent sub-questions \cite{perez2020unsupervised, xu2022decomprc}.
\paragraph{Knowledge Graphs in LLMs} LLMs have shown remarkable capabilities across a diverse range of NLP tasks, including knowledge-intensive ones like question answering and reasoning \cite{khot2022instruction}. However, integrating structured knowledge representations, such as knowledge graphs, into LLMs has been an active area of research. Studies have explored methods like knowledge injection during pre-training \cite{wang2021keytextretrieval} or graph-aware attention mechanisms \cite{zhang2022embedding}.
\paragraph{Chain of Thought Prompting} A promising approach to improve LLM reasoning is Chain of Thought (CoT) prompting \cite{wei2022chain}, which involves prompting the model to generate an explicit step-by-step reasoning process before producing the final output. The coT has shown performance improvements in reasoning tasks like arithmetic word problems \cite{wei2022chain} and commonsense question answering \cite{dasgupta2022language}, but its application to question decomposition is largely unexplored.
\paragraph{Faithfulness in LLMs} As LLMs become more prevalent, ensuring their faithfulness (adherence to factual and logical constraints) has emerged as a critical concern \cite{lin2021truthfulqa, zhang2022faithful}. 
In the context of question decomposition for large language models (LLMs), ensuring that generated sub-questions remain faithful to the original intent and logic is essential. Several recent studies have explored various techniques to improve this aspect of LLM functionality:

\begin{enumerate}
  \item \textbf{Prompting Techniques}: An in-depth analysis of different prompting methods for LLMs suggests that well-structured prompts can significantly enhance the performance and applicability of these models in complex reasoning tasks. Techniques such as ``Chain-of-Thought'' prompting enable LLMs to process information sequentially, thereby maintaining logical consistency and alignment with the initial query's intent~\cite{wei2022chain}.
  
  \item \textbf{Multi-task Learning}: Studies like \cite{gao2021better} have shown that multi-task learning can be effectively leveraged to improve the few-shot learning capabilities of LLMs. By fine-tuning models on a small set of examples across various tasks, these models can generalize better and respond more accurately to diverse inputs.
  
  \item \textbf{Constrained Decoding}: Efficient prompting methods not only guide the generation process but also constrain it to generate outputs that adhere more closely to desired outcomes. This is particularly useful in tasks where maintaining the integrity of the generated content is crucial. Techniques for efficient prompting include optimizing the design of prompts and compressing them to reduce the computational burden~\cite{chang2022efficient}.
\end{enumerate}

Previous work on retrieval-augmented generation, such as SciPhyRAG \cite{anand2023sciphyrag}, highlights the benefits of incorporating relevant documents to enhance LLM performance in physics question answering. Our approach extends this concept by using knowledge graphs to maintain logical consistency in sub-question generation.

\section{Dataset Description}
In this section, we introduce a novel dataset comprising high-quality, high-school physics questions, augmented with corresponding internal knowledge graphs and subqueries generated using advanced models such as Gemini Pro. This dataset is designed to enhance the capabilities of open-source large language models (LLMs) by providing them with structured reasoning paths. These structured reasoning paths containing query's knowledge graph and model-generated subqueries within our dataset offer a unique resource for fine-tuning open-source LLMs, aiming to replicate and expand upon the sophisticated analytical abilities demonstrated by larger models like GPT-4 and Gemini Pro. This enhancement mainly focuses on educational applications, leveraging knowledge graph-based query decomposition to improve LLMs reasoning and performance.

\subsection{Physics Question Bank}
For collecting physics questions, we utilize the dataset proposed in \cite{anand2023revolutionizing}. This dataset comprises approximately 8,000 meticulously augmented high school-level physics questions. It covers a wide range of topics, including mechanics, electromagnetism, thermodynamics, optics, and atomic physics. This comprehensive dataset is designed to rigorously test the reasoning capabilities of large language models (LLMs) in educational contexts, focusing on these core areas of physics. We use this dataset as our physics question bank. 

\subsection{Knowledge Graphs \& Subqueries}
For a given question $Q$, we first instruct Gemini Pro to generate a knowledge graph $K$ using all the information provided in $Q$, allowing the internal logic and relationships within the query to be captured in $K$. Next, the LLM is directed to derive subqueries from $Q$ based on the structure and content of $K$. These subqueries are solved independently by the LLM. The solutions to these subqueries are then used to construct a comprehensive response for $Q$. This method ensures that each component of the original question is addressed thoroughly. The complete response is then collected to parse the knowledge graph and subqueries for each question in our physics question bank. The collected knowledge graphs and subqueries establish the reasoning path for each question. Using this method, we obtain a rich dataset of 8000 questions and their corresponding reasoning path (knowledge graphs + subqueries).

\subsection{Dataset Sample}
Each data point in our dataset comprises three integral components: Question, Knowledge Graph, and Subqueries. Below is an example to illustrate this structure:
\subsubsection{Question}
"Calculate the stress developed inside a tooth cavity filled with copper when hot tea at a temperature of 26.65 °C is drunk. Assume the body temperature to be 37 °C and the coefficient of linear expansion, \(\alpha=6.74 \times 10^{-6}\) °C\(^{-1}\). The bulk modulus for copper is \(6.28 \times 10^9\) N/m\(^{2}\)."

\subsubsection{Knowledge Graph} 
\begin{verbatim}
{
 "nodes": [
  {
   "id": "tooth_cavity",
   "properties": {
   "material": "copper",
   "temperature_initial": "26.65 °C",
   "temperature_final": "37 °C",
   "coefficient_of_expansion": "6.74e-06 
   °C^-1"
    }
  },
  {
   "id": "bulk_modulus",
   "properties": {
   "value": "6.28e+09 N/m^2"
   }
  }
  ],
  "edges": [
   {
    "source": "tooth_cavity",
    "target": "bulk_modulus",
    "label": "has_bulk_modulus"
   }
  ]
}
\end{verbatim}

\subsubsection{Subqueries}
Subquery 1: Calculate the change in temperature of the tooth cavity. \\
Subquery 2: Calculate the strain experienced by the tooth cavity. \\
Subquery 3: Calculate the stress generated within the tooth cavity.

\section{Experimentation and Methodology}
\paragraph{Generation of Knowledge Graphs}
To conduct this study, we began by creating Knowledge Graphs utilizing large language models, which are known for their strength in tasks such as entity and relationship detection. 

These Knowledge graphs were then utilized to guide a model in the creation of Sub-Queries, as is common in Task Decomposition. Sub-queries are essential for dividing complex questions into manageable parts that can be individually addressed.

\paragraph{Sub-Query Processing}
Each Sub-Query is passed to an answering module, which constructs a few-shot prompt for the given Sub-Query to improve Sub-Question response accuracy.

\paragraph{Synthesis Module}
The responses for the Sub-Queries are further passed to a synthesis module, along with the original question. The synthesis module is then prompted with the original question, along with the responses for the Sub-Queries given as facts.

This pipeline serves to improve response quality for Question-Answering tasks, along with increasing Model Faithfulness and Transparency in Answering. This is particularly beneficial in enhancing the utility of the system for educational purposes.

\begin{figure}[h!]
  \centering
  \includegraphics[width=0.5\textwidth]{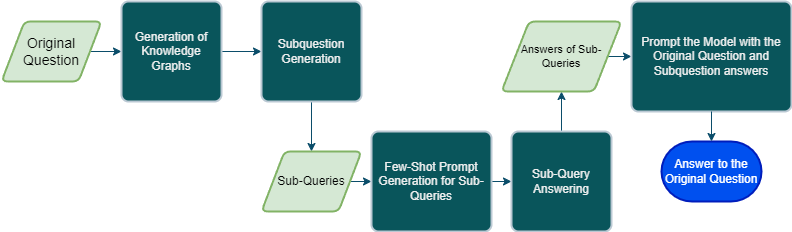}
  \caption{Overview of the question-answering pipeline. The process begins with converting the original question into a Knowledge Graph to identify and relate various entities. Sub-questions are then generated based on this graph. After answering these sub-questions, the model is reprompted with the original question, incorporating the answers to sub-questions as factual inputs to enhance the accuracy and relevance of the final response.}
  \label{fig:image1}
\end{figure}
To analyze the effectiveness of our approach, we utilized the novel dataset introduced in the 'Revolutionizing High School Physics Education' study\cite{anand2023revolutionizing}. The dataset's structure allows for a comprehensive assessment of the model's ability to generate relevant sub-questions.

The MM-PhyQA model's multimodal approach, incorporating images and chain-of-thought prompting, offers a useful comparison for our method's performance in handling complex physics questions \cite{anand2024mm}. Additionally, the Mathify study provides insights into evaluating LLMs on structured problem-solving tasks, which is analogous to our assessment of sub-question generation \cite{anand2024mathify}. 

\section{Analysis}
\subsection{Human Evaluation}
In this section, we present a comprehensive human evaluation conducted using a set of 100 unique questions sourced from our physics question bank. The objective is to assess the performance enhancements achieved by our proposed methodology by comparing them with baseline reasoning methods. We chose GPT-4 for this comprehensive evaluation. The reasoning methods include: 

\begin{enumerate}
\item \textbf{Standard Prompting}: This refers to directly inputting the questions into GPT-4 without any preparatory modifications or instructions within the prompt.

\item \textbf{Question Decomposition without Knowledge Graph}: In this method, GPT-4 is instructed to break down each question into smaller, manageable sub-queries, resolve these independently, and synthesize the responses to formulate the final answer.

\item \textbf{Question Decomposition with Knowledge Graph}: This approach employs our novel method, which first directs GPT-4 to construct a knowledge graph from the information embedded in the question. Subsequently, GPT-4 generates sub-queries based on this knowledge graph and utilizes the responses to these sub-queries to answer the original question comprehensively.
\end{enumerate}
The chosen subset of questions from the question bank can be divided into two broad categories:
\begin{enumerate}
    \item \textbf{Numerical-Based Solving:} This category encompasses all numerically-based questions within our dataset that require the application of specific formulas for correct resolution. The questions vary, including those that necessitate a single formula and those requiring the integration of multiple formulas within a specific topic of interest.\\

    \textbf{Example:} "A piece of wood from the ruins of an ancient building was found to exhibit a \(^{14}C\) activity of 12 disintegrations per minute per gram of its carbon content. The \(^{14}C\) activity of living wood is typically 16 disintegrations per minute per gram. Calculate the time elapsed since the tree, from which the wooden sample was derived, died. Assume the half-life of \(^{14}C\) is 5760 years."
    
    \item \textbf{Conceptual Reasoning:} This category includes theory-based questions that primarily require reasoning with concepts from the topic of interest. These questions do not involve numerical computations but instead demand a thorough theoretical understanding and the ability to reason conceptually.\\

    \textbf{Example:} "Consider a current-carrying circular loop of radius \(R\) placed in the \(xy\)-plane with its center at the origin. Half of the loop on the side where \(x > 0\) is now bent so that it lies in the \(yz\)-plane."

    "Input:\\
     A. The magnitude of the magnetic moment now diminishes.\\
     B. The magnetic moment does not change.\\
     C. The magnitude of \(B\) at \((0,0,z)\), where \(z \gg R\), increases.\\
     D. The magnitude of \(B\) at \((0,0,z)\), where \(z > R\), is unchanged.\\"

\end{enumerate}
We evaluated all three methods mentioned above across both categories, observing the reasoning path taken with all three methods and performing the manual qualitative evaluation. The success rate measured across both the categories using all three methods are depicted in Table \ref{tab:successrate}.
\subsubsection{Numerical-Based Solving}
For numerical-based solving, all three methodologies followed similar reasoning paths, often leading to correct final answers. However, some complex questions necessitated a deep understanding of underlying concepts beyond straightforward formulaic applications. In these instances, the approaches involving standard prompting and question decomposition without a knowledge graph sometimes led GPT-4 to apply incorrect concepts, resulting in erroneous solutions. Conversely, using the knowledge graph for the question decomposition enhanced the model's capability to adhere to pertinent concepts. This method facilitated the generation of relevant subqueries, thereby guiding GPT-4 toward the correct final answer.
\subsubsection{Conceptual Reasoning}
Questions requiring conceptual reasoning displayed interesting trends. In scenarios where direct use of simple formulas was required for reasoning, all three methodologies performed well, yielding correct answers. However, for more abstract conceptual questions, GPT-4 often struggled to provide pertinent reasoning and occasionally generated incorrect responses, resulting in hallucinations. While the methods successfully generated relevant subqueries, GPT-4 frequently failed to accurately resolve these subqueries, highlighting the potential need for external knowledge sources to enhance reasoning in such contexts.\\

Notably, question decomposition with knowledge graph contributed to improvements in performance. Similar to its impact on numerical-based solving, the knowledge graph aided GPT-4 in adhering to relevant concepts, thus preventing hallucination and leading to more accurate answers. However, overall performance was less pronounced than those observed in numerical-based solving, primarily due to the model's limited internal knowledge of the subqueries. This observation underscores the need to enhance LLMs' internal knowledge base further to bolster their reasoning capabilities in conceptual domains.\\

\begin{table}[htbp]
\hspace*{-0.5 em}
\begin{tabular}{|p{0.19\textwidth}|p{0.1\textwidth}|p{0.14\textwidth}|}
\hline
Method & Numerical Solving & Conceptual Reasoning\\ 
\hline
Standard Prompting & 77.62 & 68.53\\ 
\hline
Decomposition without KG & 81.33 & 71.30\\ 
\hline
Decomposition with KG & 84.45 & 74.58\\ 
\hline
\end{tabular}
\caption{Human Evaluation of Different Reasoning Methods with GPT-4 across a chosen set of 100 questions, including 54 numerical solving and 46 conceptual reasoning questions measured by Success Rate.}
\label{tab:successrate}
\end{table}

\subsubsection{Example}
Here's an example that demonstrates how the performance of GPT-4 on a conceptual reasoning question is enhanced through question decomposition with a knowledge graph.\\ \\
\textbf{Question:} "One morning, Ramesh began to fill a bucket with hot water from a geyser to take a bath, filling it to \(\frac{1}{3}\) of its capacity. The remaining \(\frac{2}{3}\) of the bucket was intended to be filled with cold water to adjust the water to a comfortable temperature. However, Ramesh was suddenly required to attend to an urgent matter, which would take approximately 5-10 minutes. Now, he faces two options:
\begin{enumerate}
    \item Fill the remaining \(\frac{2}{3}\) of the bucket with cold water immediately, then attend to the urgent matter.
    \item Attend to the urgent matter first and fill the remaining \(\frac{2}{3}\) of the bucket with cold water just before taking the bath."
\end{enumerate}
"Which option do you think would have kept the water warmer? Explain."\\ \\
\textbf{Ground Truth:} The first option will keep the water warmer.\\ \\
\textbf{Standard Prompting:}
\begin{verbatim}
Heat Loss Considerations

The key factor here is the rate and amount of 
heat loss, which are influenced by:
1.Temperature difference
2.Surface area exposed
3.Water volume

Conclusion
Scenario 2 (attending to the work first and 
then filling the remaining bucket) is likely 
to keep the water warmer compared to 
filling it all up at once and then 
letting it sit. In Scenario 2, the hot 
water has less time to cool down in the 
mixed state with a greater volume of 
cold water, hence retaining more of its 
initial heat when finally mixed.
\end{verbatim}
\textbf{Question Decomposition without Knowledge Graph:} 
\begin{verbatim}
Subquery Decomposition 
• What happens to the temperature of the 
  hot water when mixed immediately with 
  cold water, followed by a wait time of 
  5-10 minutes? 
• What happens to the temperature of the 
  hot water if it sits alone for 5-10 
  minutes and is then mixed with cold 
  water just before use? 

Conclusion and Final Answer
The option that would have kept the water 
warmer is to first attend to the work and 
fill the remaining bucket just before 
taking the bath. This approach minimizes 
the total time that the warmer mixture is 
exposed to the cooler ambient conditions, 
thereby retaining more heat. 
\end{verbatim}
\textbf{Question Decomposition with Knowledge Graph:} 
\begin{verbatim}
Knowledge Graph in JSON Format:  

{
"nodes": {
"Ramesh": {"type": "person"},
"Bucket": {"type": "container", 
"initial_fill": "hot water", 
"remaining_fill": "cold water"},
"Hot Water":{"amount": "1/3 bucket"},
"Cold Water": {"amount": "2/3 bucket"},
"Task": {"duration": "5-10 minutes"}
},
"edges": [
{"from": "Hot Water", "to": "Bucket", 
"relationship": "fills"},
{"from": "Cold Water", "to": "Bucket", 
"relationship": "intended_to_fill"},
{"from": "Task", "to": "Ramesh", 
"relationship": "needs_attention"}
]
}

Creating Subqueries
1. How does the initial temperature of the 
   hot water change over time if left in 
   the bucket? 
2. What is the thermal impact of adding cold 
   water immediately vs. after the delay? 

Final Answer
Given the nature of heat loss and the 
properties of water mixing, filling the 
bucket completely with cold water and then 
attending to the work is likely to keep 
the water warmer by the time Ramesh returns. 
This is because the larger volume of mixed
water retains heat better over time than the 
smaller volume of just hot water cooling down 
alone. Hence, this option is more effective in 
keeping the water warm until Ramesh is ready 
to take his bath. 
\end{verbatim}
\textbf{Analysis:} Generating relevant subqueries with the knowledge graph enhances the rationale, leading to a scientifically correct answer, as opposed to the wrong answer obtained with standard prompting or question decomposition without KG.

\section{Student Survey}
To assess the effectiveness of our proposed knowledge graph-based question decomposition approach and its impact on the learning experience, we conducted a survey involving five high school students. The survey was designed to compare the output quality of different methods and quantify the students' preference for each method.
\subsection{Survey Design}
The survey consisted of the following components:
\begin{enumerate}
\item A question designed to gauge participants' familiarity with high school physics, rated on a scale from 1 (Weak) to 5 (Strong).
\item Three sets of questions, each containing:
\begin{itemize}
\item The original physics question
\item Sub-questions generated using three different methods:
\begin{enumerate}
\item Standard Prompting
\item Question decomposition without knowledge graphs
\item Question decomposition with knowledge graphs
\end{enumerate}
\item A rating scale from 1 to 5 for each method, assessing the clarity, logical consistency, and helpfulness of the sub-questions in solving the original question.
\end{itemize}
\item An open-ended question asking participants to provide feedback on their preferred method and the reasons behind their choice.
\end{enumerate}
\subsection{Survey Findings}
\begin{enumerate}
\item Familiarity with High School Physics: All five participants rated their familiarity with high school physics between 4 and 5, indicating a strong grasp of the subject.
\item Output Quality and Preference:
\begin{itemize}
\item For the first set of questions, four out of five students rated the knowledge graph-based decomposition method the highest (average rating: 4.6), citing improved logical consistency and relevance of the sub-questions to the original question.
\item In the second set, all five students preferred the knowledge graph-based approach (average rating: 4.8), highlighting the clarity and contextual appropriateness of the generated sub-questions.
\item For the third set, three students favored the knowledge graph-based method (average rating: 4.4), while two students found the question decomposition without knowledge graphs to be equally effective (average rating: 4.2).
\end{itemize}

\begin{figure}[h!]
  \centering
  \includegraphics[width=0.5\textwidth]{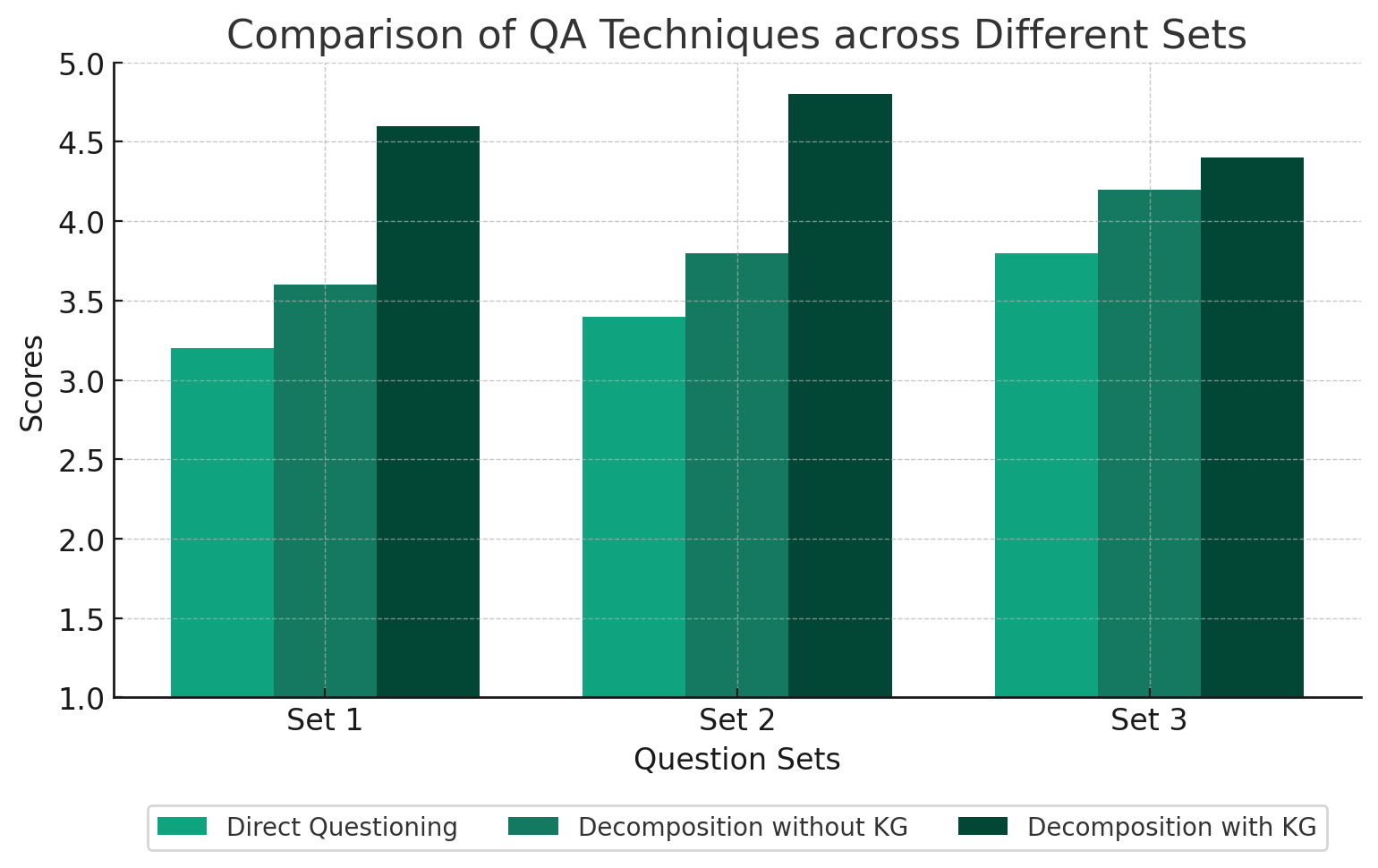}
  \caption{Bar chart showing the performance of three QA techniques across different question sets, highlighting the effectiveness of each method with scores from 1 to 5.}
  \label{fig:image2}
\end{figure}

\item Feedback: Participants appreciated the structured and guided approach of the knowledge graph-based question decomposition, stating that it helped them better understand the problem-solving process and enhanced their learning experience. They also mentioned that this method could be particularly useful for complex, multi-step questions.
\end{enumerate}
\subsection{Implications}
The survey results suggest that the knowledge graph-based question decomposition approach is well-received by high school students, with the majority preferring this method over traditional questioning techniques. The findings support our hypothesis that this approach leads to more logically consistent and relevant sub-questions, ultimately improving the learning experience and problem-solving process.
However, given the limited sample size of this survey, further research with a larger and more diverse group of participants is necessary to generalize these findings. Additionally, future studies could explore the effectiveness of this approach across different subjects and difficulty levels to assess its broad applicability in educational contexts.

\section{Limitations}
\paragraph{}While the ability to discretize the operations inherent to a question-answering task does lend itself to higher-quality answers, this atomization of the task results in potentially increased inference costs, due to potential redundant information processing. All experimentation was done in the absence of Fine-tuning due to difficulty in synthesizing large amounts of training data, which presents an opportunity for improvement should such a hurdle be overcome.
\paragraph{}The results presented in this study are domain-specific; while the results may be generalized to similar fields, such as Mathematics, it may be difficult to extrapolate such performance gains to more diverse domains. Quantitatively evaluating such a pipeline is quite different in the absence of large amounts of pre-existing data and a large team of domain experts. While we have noticed significant qualitative improvements, future studies may be well served by a well-framed quantitative metric that can precisely capture the relevant features of the responses.
\section{Discussion}
Our findings indicate that sub-questions derived from knowledge graphs exhibit significantly improved fidelity to the original questions' logic. This improvement not only enhances the learning experience by providing clearer, contextually appropriate sub-questions but also demonstrates the potential of LLMs to revolutionize educational methodologies. 


\section{Conclusion}
Our study demonstrates that leveraging knowledge graphs for question decomposition in LLMs can significantly enhance the quality of generated subquestions and overall model performance. This is consistent with previous research showing the benefits of structured knowledge integration in AI systems \cite{anand2023revolutionizing}. 
The experimental results demonstrate that our knowledge graph-based question decomposition approach leads to significant improvements in the quality and faithfulness of the generated sub-questions compared to traditional decomposition techniques. This, in turn, results in higher-quality answers that are more aligned with the original question's logic and structure.

The proposed methodology has several implications for the application of LLMs in educational settings. By providing clearer and more contextually appropriate sub-questions, our approach can enhance the learning experience for students, allowing them to better understand complex concepts and problem-solving strategies. Moreover, the improved performance of LLMs in question-answering tasks highlights their potential to revolutionize educational methodologies and support personalized learning.

However, it is important to acknowledge the limitations of this study. While we focused on the domain of high school-level physics, further research is needed to investigate the generalizability of our approach to other educational domains and question types. Additionally, the scalability of the proposed pipeline and its integration with existing educational platforms require further exploration.

Future work should focus on extending the proposed methodology to a wider range of educational contexts and investigating the long-term impact of LLM-based question answering on student learning outcomes. Moreover, the development of more advanced knowledge graph construction techniques and the incorporation of external knowledge sources could further enhance the performance and versatility of our approach.

In conclusion, this study demonstrates the potential of knowledge graph-based question decomposition in improving the question-answering capabilities of LLMs for high school-level physics education. By bridging the gap between structured knowledge representations and the reasoning capabilities of LLMs, our approach opens up new avenues for the application of artificial intelligence in education, paving the way for more effective and personalized learning experiences.

\begin{appendix}

\section{Appendix: Additional Experimentation Details}

\subsection{Example from Experimentation}

To provide further clarity on the methodology and results, we present an example question from the physics question bank along with the corresponding knowledge graph and subqueries.

\subsubsection{Original Question}
"Calculate the stress developed inside a tooth cavity filled with copper when hot tea at a temperature of 26.65 °C is drunk. Assume the body temperature to be 37 °C and the coefficient of linear expansion, \(\alpha=6.74 \times 10^{-6}\) °C\(^{-1}\). The bulk modulus for copper is \(6.28 \times 10^9\) N/m\(^{2}\)."

\subsubsection{Knowledge Graph}
\begin{verbatim}
{
 "nodes": [
  {
   "id": "tooth_cavity",
   "properties": {
   "material": "copper",
   "temperature_initial": "26.65 °C",
   "temperature_final": "37 °C",
   "coefficient_of_expansion": "6.74e-06 
   °C^-1"
    }
  },
  {
   "id": "bulk_modulus",
   "properties": {
   "value": "6.28e+09 N/m^2"
   }
  }
  ],
  "edges": [
   {
    "source": "tooth_cavity",
    "target": "bulk_modulus",
    "label": "has_bulk_modulus"
   }
  ]
}
\end{verbatim}

\subsubsection{Subqueries}
1. Calculate the change in temperature of the tooth cavity.
2. Calculate the strain experienced by the tooth cavity.
3. Calculate the stress developed inside the tooth cavity.

\subsubsection{Subquery Answers and Synthesis}
1. \textbf{Change in Temperature}:
\begin{equation}
   \Delta T = T_{\text{final}} - T_{\text{initial}} = 37 - 26.65 = 10.35 \, \text{°C}
\end{equation}
2. \textbf{Strain}: $\alpha \cdot \Delta T =$\\
   \begin{equation}
   6.74 \times 10^{-6} \, \text{°C}^{-1} \times 10.35 \, \text{°C} = 6.98 \times 10^{-5}
   \end{equation}
3. \textbf{Stress}: $\text{Strain} \times \text{Bulk Modulus} =$\\
\begin{equation}
   6.98 \times 10^{-5} \times 6.28 \times 10^9 \, \text{N/m}^2 = 438,344 \, \text{N/m}^2
\end{equation}

\subsubsection{Final Answer}
The stress developed inside the tooth cavity filled with copper is approximately \(438,344 \, \text{N/m}^2\).

\subsection{Survey Results Details}
The survey conducted among five high school students provided insights into the effectiveness of different question decomposition methods. Here are additional details:

\subsubsection{Survey Questions}
1. Rate your familiarity with high school physics on a scale from 1 (Weak) to 5 (Strong).
2. Evaluate the clarity, logical consistency, and helpfulness of sub-questions generated by:
   a. Standard Prompting
   b. Question Decomposition without Knowledge Graphs
   c. Question Decomposition with Knowledge Graphs

\subsubsection{Detailed Results}
The average ratings (on a scale of 1 to 5) for each method were as follows:

\begin{table}[h]
\centering
\begin{tabular}{{|p{0.19\textwidth}|p{0.1\textwidth}|p{0.14\textwidth}|}}
\hline
Method & Numerical Solving & Conceptual Reasoning \\
\hline
Standard Prompting & 3.8 & 3.4 \\
\hline
Decomposition without KG & 4.1 & 3.6 \\
\hline
Decomposition with KG & 4.5 & 4.2 \\
\hline
\end{tabular}
\caption{Average Ratings from Student Survey}
\end{table}

\subsubsection{Student Feedback Highlights}
- \textbf{Student A}: "The knowledge graph-based method made it easier to follow the logical steps needed to solve the problem." \\
- \textbf{Student B}: "I appreciated the structured approach, which helped me understand the connections between different parts of the question."
\end{appendix}
\bibliography{acl_latex}




\end{document}